\documentclass[sigconf]{acmart}

\usepackage{graphicx}
\usepackage{booktabs}
\usepackage{subcaption}
\usepackage{multirow}
\usepackage{bm}
\usepackage{colortbl}
\usepackage{listings}
\usepackage{float} 
\usepackage{nicematrix}  

\definecolor{ACMHighlight}{RGB}{230,230,230}
\definecolor{lightblue}{RGB}{240,248,255}
\definecolor{lightgreen}{RGB}{240,255,240}

\AtBeginDocument{%
  }

\copyrightyear{2026}
\acmYear{2026}
\setcopyright{cc}
\setcctype{by}
\acmConference[WWW '26] {Proceedings of the ACM Web Conference 2026}{April 13--17, 2026}{Dubai, United Arab Emirates.}
\acmBooktitle{Proceedings of the ACM Web Conference 2026 (WWW '26), April 13--17, 2026, Dubai, United Arab Emirates}
\acmISBN{979-8-4007-2307-0/2026/04}
\acmDOI{xx.xxxx/xxxxxxxx.xxxxxxx}

\begin{document}

\title{Beyond Words: Enhancing Desire, Emotion, and Sentiment Recognition with Non-Verbal Cues}


\author{Wei Chen}
\orcid{0009-0000-7820-3498}
\affiliation{%
	\institution{College of Informatics, Huazhong Agricultural University}
	\city{Wuhan}
	\country{China}}
\email{weichen5498@webmail.hzau.edu.cn}

\author{Tongguan Wang}
\orcid{0009-0001-8747-1790}
\affiliation{%
	\institution{College of Informatics, Huazhong Agricultural University}
	\city{Wuhan}
	\country{China}}
\email{wang\_tg@webmail.hzau.edu.cn}

\author{Feiyue Xue}
\orcid{0009-0007-0949-2987}
\affiliation{%
	\institution{College of Informatics, Huazhong Agricultural University}
	\city{Wuhan}
	\country{China}}
\email{xuefeiyue@webmail.hzau.edu.cn}

\author{Junkai Li}
\orcid{0009-0001-3236-6838}
\affiliation{%
	\institution{College of Informatics, Huazhong Agricultural University}
	\city{Wuhan}
	\country{China}}
\email{junkaili@webmail.hzau.edu.cn}

\author{Hui Liu}
\orcid{0009-0003-9920-1186}
\affiliation{%
	\institution{College of Informatics, Huazhong Agricultural University}
	\city{Wuhan}
	\country{China}}
\email{liuhui\_1003@webmail.hzau.edu.cn}

\author{Ying Sha}
\orcid{0000-0002-6638-5009}
\authornote{Corresponding author.}
\authornote{Also with Hubei Engineering Technology Research Center of Agricultural Big Data, Key Laboratory of Smart Farming for Agricultural Animals.}
\affiliation{%
	\institution{College of Informatics, Huazhong Agricultural University}
	\institution{Engineering Research Center of Intelligent Technology for Agriculture}
	\city{Wuhan}
	\country{China}} 
\email{shaying@mail.hzau.edu.cn}

\renewcommand{\shortauthors}{Wei Chen et al.}

\begin{abstract}	
	
	Multimodal desire understanding, a task closely related to both emotion and sentiment that aims to infer human intentions from visual and textual cues, is an emerging yet underexplored task in affective computing with applications in social media analysis. Existing methods for related tasks predominantly focus on mining verbal cues, often overlooking the effective utilization of non-verbal cues embedded in images. To bridge this gap, we propose a \textit{\textbf{Sy}mmetrical Bidirectional Multimodal Learning Framework for \textbf{D}esire, \textbf{E}motion, and \textbf{S}entiment Recognition} (SyDES). The core of SyDES is to achieve bidirectional fine-grained modal alignment between text and image modalities. Specifically, we introduce a mixed-scaled image strategy that combines global context from low-resolution images with fine-grained local features via masked image modeling (MIM) on high-resolution sub-images, effectively capturing intention-related visual representations. Then, we devise symmetrical cross-modal decoders, including a text-guided image decoder and an image-guided text decoder, which enable mutual reconstruction and refinement between modalities, facilitating deep cross-modal interaction. Furthermore, a set of dedicated loss functions is designed to harmonize potential conflicts between the MIM and modal alignment objectives during optimization. Extensive evaluations on the MSED benchmark demonstrate the superiority of our approach, which establishes a new state-of-the-art performance with 1.1\% F1-score improvement in desire understanding. Consistent gains in emotion and sentiment recognition further validate its generalization ability and the necessity of utilizing non-verbal cues. Our code is available at: \href{https://github.com/especiallyW/SyDES}{https://github.com/especiallyW/SyDES}.
		
\end{abstract}



\maketitle

\section{Introduction}\label{sec1}

\begin{figure}[tbp]
		\centering
		\includegraphics[width=\columnwidth]{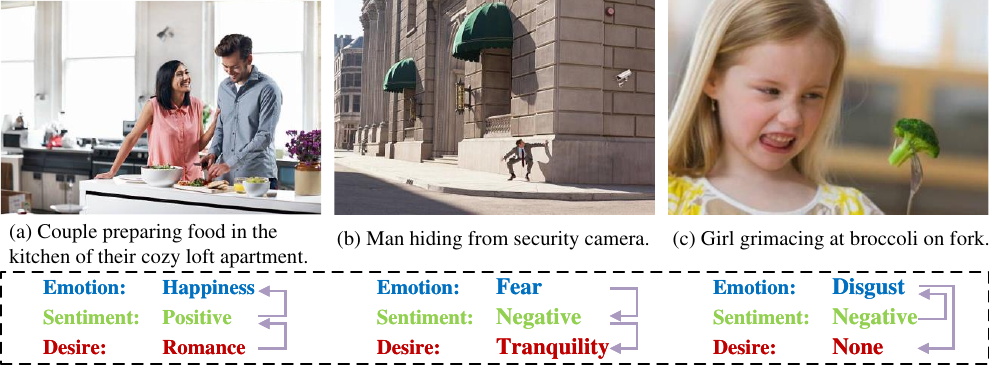}
		\caption{Examples of multimodal desire, emotion, and sentiment.}
		\label{fig:dataset}
	\end{figure}

	Human Desire is a fundamental intention that reflects a strong wish for certain objects or states \cite{Portner2020DesireBA}. It interacts closely with emotion and sentiment, shaping affective life experiences while being modulated by them in return. Such three tasks form interconnected and essential components of the human experience, driving our actions and decisions. For example, in Figure~\ref{fig:dataset} (a), a couple smiling and preparing food in a kitchen can be interpreted as expressing a "\textit{\textbf{romance}}" desire, which explains positive sentiment and happy emotion. Figure~\ref{fig:dataset} (b) depicts a man’s exaggerated movements to avoid security cameras. It can be understood as fear and negative sentiment driven by a desire for safety and privacy. Therefore, if a machine were capable of accurately inferring such desire intents, it would move research closer to recognizing human emotional intelligence \cite{hofmann2015psychology}. However, the specific problem of desire understanding has received comparatively little dedicated attention.
	
	A key to desire understanding lies in effectively leveraging the non-verbal cues embedded in images, which provide rich contextual information beyond textual descriptions. However, existing methods in multimodal emotion and sentiment recognition \cite{Yang2021MultimodalSD, Zhu2023MultimodalEC} still predominantly focus on enhancing verbal cues (e.g., via graph neural networks), treating image-based non-verbal cues merely as auxiliary features to be extracted and fused superficially. This underutilization constitutes a fundamental limitation. In practice, non-verbal cues play a crucial role. For instance, the grimacing expression in Figure~\ref{fig:dataset} (c) could indicate disgust toward broccoli or be part of a playful interaction with family; without the rich context from the image, text-based inference is inherently ambiguous. We argue that overcoming this limitation is not only beneficial for emotion and sentiment analysis but is particularly critical for advancing the more complex task of desire understanding.
	
	To this end, we propose SyDES, \textit{a \textbf{Sy}mmetrical Bidirectional Multimodal Learning Framework for \textbf{D}esire, \textbf{E}motion, and \textbf{S}entiment recognition}. The framework emphasizes deep utilization of non-verbal visual cues while ensuring that verbal cues remain effectively exploited. Specifically, the input image is processed as both a low-resolution version and a high-resolution version using a shared image encoder. The low-resolution image provides global visual representations for cross-modal alignment. The high-resolution image is processed by mixed-scale image strategy to get high-resolution sub-images, and these sub-images are modeled with masked image modeling to encourage the encoder to better learn fine-grained local features. A text-guided image decoder is introduced so that image reconstruction can be guided by textual semantics. Conversely, an image-guided text decoder is employed so that text decoding can incorporate multi-scale visual information. In the meantime, we design a set of loss functions (e.g., local-global semantic similarity loss, cross-modal feature-distribution consistency loss) to allow reconstructed image to maintain modal alignment of fine-grained local visual features and global visual representations, thereby avoid over-reliance on specific regional features. These mechanisms enable mutual guidance and semantic alignment of textual representations, local visual features and global visual features. The fused text outputs are then passed to a lightweight multi-layer perception (MLP) for downstream prediction (see Figure~\ref{fig:model} and Section~\ref{sec3} for details).
	
	To evaluate our approach, extensive experiments are conducted on MSED dataset, the first multimodal benchmark encompassing desire understanding, emotion recognition, and sentiment analysis. Our approach is primarily evaluated on the desire understanding task, where it achieves a significant improvement of 1.1\% in F1-score, establishing a new state-of-the-art. Furthermore, consistent performance gains of 0.6\% in emotion recognition and 0.9\% in sentiment analysis demonstrate the framework's generalizability and underscore the necessity of fully utilizing non-verbal cues. Our contributions can be summarized: 
	
	\begin{enumerate}
		\item We propose SyDES, a novel framework by deeply leveraging non-verbal visual cues through a symmetrical bidirectional architecture.
		\item We introduce a mixed-scale image strategy and symmetrical cross-modal decoders, to capture fine-grained features and deep cross-modal interaction between text and image modalities.
		\item We design a set of dedicated loss functions to harmonize the objectives of MIM and modal alignment, ensuring consistent learning across different modalities and scales.
		\item We provide comprehensive experiments and ablation studies on the MSED dataset, validating the effectiveness and generalization of our proposed SyDES for multimodal desire understanding, emotion recognition, and sentiment analysis.
	\end{enumerate}
	
\section{Related Work}\label{sec2}

	\subsection{Sentiment Analysis and Emotion Recognition} \label{subsec21}
	
	\subsubsection{Text-only sentiment analysis} \label{subsec211}
	
		Sentiment analysis \cite{Kim2014ConvolutionalNN, pang2002thumbs, Tai2015ImprovedSR, Singh2021SentimentAO} has long been a central topic in natural language processing. Early studies were largely unimodal. For example, Taboada et al. \cite{taboada2011lexicon} proposed the lexicon-based Semantic Orientation CALculator, which leverages polarity-annotated lexica with negation handling. Pang et al. \cite{pang2002thumbs} and Gamallo et al. \cite{gamallo2014citius} applied classical classifiers (SVM, Naïve Bayes) to sentiment tasks. Kim et al. \cite{Kim2014ConvolutionalNN} introduced convolutional neural networks for text classification, while Tai et al. \cite{Tai2015ImprovedSR} proposed Tree-LSTM to model syntactic hierarchies. Yang et al. \cite{Yang2016HierarchicalAN} developed the Hierarchical Attention Network (HAN) to select salient words and sentences for document-level classification. Wang et al. \cite{Wang2016AttentionbasedLF} incorporated attention into LSTM for aspect–sentiment modeling. More recently, Singh et al. \cite{Singh2021SentimentAO} employed BERT for sentiment analysis on COVID-related tweets. These methods focus on textual signals; however, social media content often pairs text with images, and multimodal cues typically provide complementary information that improves predictive accuracy.
	
	\subsubsection{Multimodal sentiment analysis and emotion recognition} \label{subsec212}
	
		Multimodal sentiment classification \cite{You2016CrossmodalityCR, Xu2018ACN, Huang2020AttentionBasedMN} has attracted increasing interest for jointly modeling sentiment across modalities. You et al. \cite{You2016CrossmodalityCR} introduced cross-modal consistency regression (CCR) to fuse image and text features. Xu et al. \cite{Xu2017MultiSentiNetAD} leveraged scene-level visual cues with attention to identify salient textual elements. Hu et al. \cite{Hu2018MultimodalSA} investigated users' latent affective states. From the interaction perspective, Xu et al. \cite{Xu2018ACN} explored iterative image–text relations; Huang et al. \cite{Huang2020AttentionBasedMN} proposed hybrid fusion of unimodal and cross-modal representations. Li et al. \cite{Li2022CLMLFAC} used contrastive learning and augmentation to align token-level image–text features. Multimodal emotion recognition \cite{Ranganathan2016MultimodalER, Zhang2022MultimodalER, Le2023MultiLabelME}, which targets finer affective states, has also been extensively studied. Guo et al. \cite{Guo2021LDMANLM} introduced multimodal news datasets and a layout-driven network; Nemaiti et al. \cite{Nemati2019AHL} proposed a hybrid latent-space fusion method; Zhang et al. \cite{Zhang2022MultimodalER} combined manifold learning with deep convolutional networks; Xu et al. \cite{Xu2017AnalyzingMP} used image captions as semantic cues; Yang et al. \cite{Yang2020ImageTextME} employed memory and multi-view attention for integration. Despite these advances, most of the existing approaches simply leveraged holistic or local features extracted from different modalities to predict multimodal sentiments, which leads to suboptimal performance.

		The emergence of graph neural networks (GNNs) \cite{Scarselli2009TheGN, Liu2019SocialRR} enabled structured relation mining among verbal cues. Yang et al. \cite{Yang2021MultimodalSD} introduced a multi-channel GNN to capture global emotional attributes. Zhang et al. \cite{Zhang2023M3GATAM} proposed a multi-task interactive graph-attention network with local–global context modules. Wang et al. \cite{Wang2024WisdoMIM} enriched text representations with contextual world knowledge from large multimodal models. However, these approaches predominantly focus on mining verbal cues and often underutilize the rich information contained in non-verbal cues.
	
		Motived by the importance of non-verbal cues, we hope to attain fine-gained features from image contextual information. Masked image modeling paradigm introduced by He et al. \cite{He2021MaskedAA} has shown that an encoder can be encouraged, via a reconstruction-based self-supervised objective, to learn richer local representations. This insight motivates our proposal of a symmetrical bidirectional multimodal learning framework to ensure more effective exploitation of image-based non-verbal cues for desire understanding, emotion recognition, and sentiment analysis.

	\subsection{Multimodal Desire Understanding} \label{subsec22}
		
		Multimodal desire understanding concerns recognizing desires or intentions expressed in both textual and visual expression, and it remains an underexplored problem. Existing automated analyses of desire largely originate from psychology and philosophy. Lim et al. \cite{lim2012desire} developed a desire-understanding system based on four emotional states in audio and gestural cues. Cacioppo et al. \cite{cacioppo2012common} designed a multi-level kernel-density fMRI analysis to investigate differences and correlations between sexual desire and love. Schutte et al. \cite{Schutte2019AMO} conducted a meta-analysis on 2,692 participants to examine links between curiosity and creativity. Hoppe et al. \cite{Hoppe2015RecognitionOC} estimated different levels of curiosity using eye-movement data and SVM. Yavuz et al. \cite{Yavuz2019ADM} proposed a data-mining approach for desire and intent using neural networks and Bayesian networks. Chauhan et al. \cite{chauhan2020all} presented a multi-task multimodal deep attentive framework for offense, motivation, and sentiment analysis. Nevertheless, these researches commonly lack support for large-scale multimodal social media data and often do not fully exploit both visual and textual channels.
		
		The recent introduction of the MSED dataset \cite{Jia2022BeyondEA}, the first multimodal dataset for desire understanding, provides a valuable benchmark. And subsequent work like MMTF-DES \cite{Aziz2025MMTFDESAF} attempts to improve performance through model ensemble. Nevertheless, such ensemble strategies incur significant computational costs and still lack a principled approach for deep, fine-grained cross-modal interaction. To address these limitations, we propose a novel symmetrical bidirectional framework that systematically leverages non-verbal cues through a mixed-scale image strategy and symmetrical decoders, offering an efficient and effective solution for multimodal desire understanding.

\section{SyDES}
\label{sec3}

	\begin{figure*}[tp]
		\centering
		\includegraphics[width=1.00\textwidth]{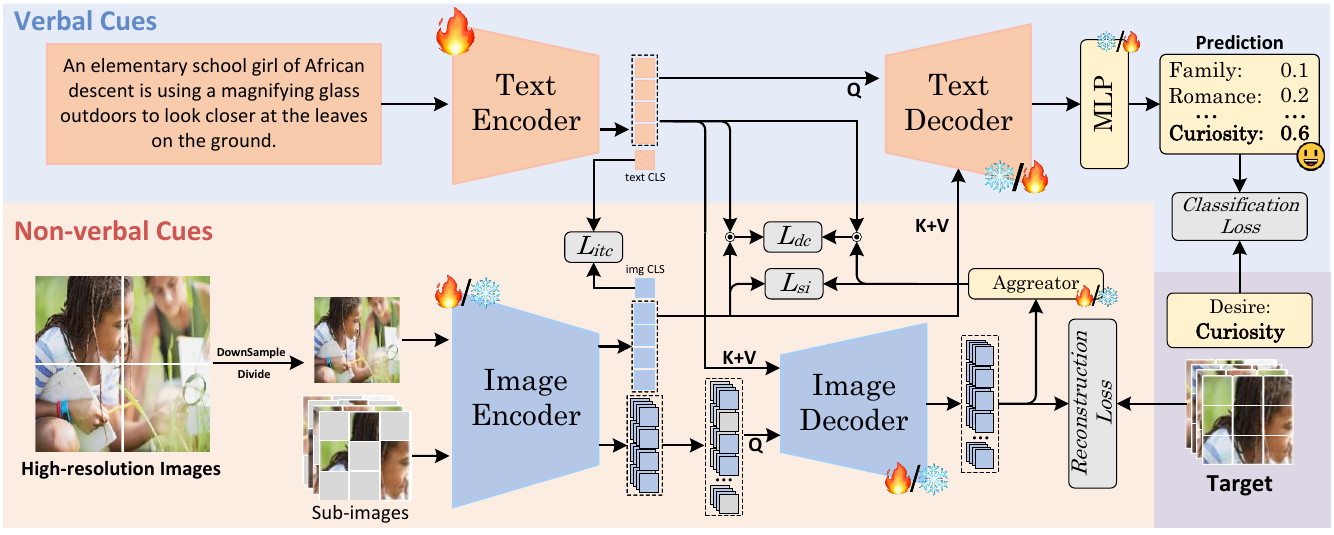}
		\caption{Overall architecture of SyDES. The model consists of four core modules: an image encoder, a text encoder, a text-guided image decoder, and an image-guided text decoder.}
		\label{fig:model}
	\end{figure*}

	Figure~\ref{fig:model} illustrates the overall architecture of SyDES. Motivated by the necessity to deeply leverage non-verbal visual cues for desire understanding, our framework is built upon a mixed-scale image strategy to capture both global context and fine-grained local features. We first detail this strategy, followed by descriptions of the image encoder, text encoder, symmetrical decoders, the loss functions designed to harmonize different learning objectives, and the two-stage training paradigm.
	
	\subsection{Mixed-Scale Image Strategy} \label{subsec31}

		A core challenge in multimodal desire understanding is to extract comprehensive intention-related representations from images, which necessitates the model's ability to perceive both global context and fine-grained local details. High-resolution images can capture richer region-level features and local details, but processing them entirely at full resolution (e.g., $448\times448$) is computationally prohibitive. Conversely, conventional low-resolution inputs (e.g., $224\times224$) sacrifice crucial local information. 

		To balance perceptual granularity and computational efficiency, we adopt a mixed-scale image strategy. Given a high-resolution image $I_i$ (e.g., $448\times448$) from batch size $N$, we generate one downsampled low-resolution image and partition the original image into four non-overlapping high-resolution sub-images. This yields five $224\times224$ images per original image, providing one global view and four detailed local views. Formally:

		\begin{equation}
			\begin{aligned}
				I_{i}^{\mathrm{(g)}} &= \operatorname{DownSample}(I_{i}), & & I_{i}^{\text{(g)}} \in \mathbb{R}^{224 \times 224 \times 3} \\ 
				I_{i,n}^{(\mathrm{l})} &= \operatorname{Crop}_{n}(I_{i}), & & I_{i,n}^{(\text{l})} \in \mathbb{R}^{224 \times 224 \times 3},\ n = 1,2,3,4
			\end{aligned}
		\end{equation}
		where $\operatorname{DownSample}$ uses bilinear interpolation, and $\operatorname{Crop}_{n}$ denotes corner or predefined cropping. Each $224\times224$ image is treated as an independent input to the image encoder to obtain its representations. This strategy efficiently provides the model with both global and local visual cues essential for understanding nuanced desires.

	\subsection{Model Architecture} \label{subsec32}
	
		Our proposed SyDES is built upon a symmetrical bidirectional architecture designed to achieve deep, bidirectional fusion between verbal and non-verbal cues. As illustrated in Figure~\ref{fig:model}, the framework comprises four core modules: image encoder, text encoder, text-guided image decoder, and image-guided text decoder, plus a lightweight MLP for downstream prediction.

		\subsubsection{Image Encoder} \label{subsec321}	
	
			We employ a vision transformer \cite{Dosovitskiy2020AnII} as the shared image encoder to process multi-scale visual representations. For the global context, we use the low-resolution image $I_{i}^{(\mathrm{g})}$, while for local details, we process high-resolution sub-images $I_{i,n}^{(\mathrm{l})}$ through masked image modeling. Specifically, each $224\times224$ image is divided into $P$ patches and embedded as:
			\[
			\begin{aligned}
				P_{\mathrm{emb},i,n}^{(\mathrm{g})} &= \left[ v_{i}^{\mathrm{cls}}, v_{i}^{1}, \ldots, v_{i}^{P} \right] \in \mathbb{R}^{(P+1) \times C_{1}} \\
				P_{\mathrm{emb},i,n}^{(\mathrm{l})} &= \left[ v_{i,n}^{\mathrm{cls}}, v_{i,n}^{1}, \ldots, v_{i,n}^{P} \right] \in \mathbb{R}^{(P+1) \times C_{1}}
			\end{aligned}
			\]
			where $C_1$ denotes the image embedding dimension and $v^{\mathrm{cls}}$ is the CLS token. For high-resolution sub-images $I_{i,n}^{(\mathrm{l})}$, we apply a binary mask vector $m_{i,n} \in \{0,1\}^{P}$ with $m_{i,n}[P]=1$ indicating that patch $p$ is masked. Following \cite{He2021MaskedAA}, we set the mask ratio to $m \in [0,1]$ and set the keep ratio to $r=1-m$. The set of unmasked indices is $M_{i,n} = \{p \mid m_{i,n}[p]=0\}$ with $|M_{i,n}| = rP$. Selecting the unmasked tokens yields:
			\[
			\widetilde{P}_{\mathrm{emb},i,n}^{(\mathrm{l})} = \mathrm{Select}\left( P_{\mathrm{emb},i,n}^{(\mathrm{l})}, M_{i,n} \right) \in \mathbb{R}^{(rP+1) \times C_{1}}
			\]
			
			Both $P_{\mathrm{emb},i}^{(\mathrm{g})}$ and $\widetilde{P}_{\mathrm{emb},i,n}^{(\mathrm{l})}$ are fed into the image encoder to produce $V_{i}^{(\mathrm{g})} \in \mathbb{R}^{(P+1) \times C_{1}}$ and $V_{i,n}^{(\mathrm{l})} \in \mathbb{R}^{(rP+1) \times C_{1}}$, where the first row of $V_{i}^{(\mathrm{g})}$ corresponds to the low-resolution CLS visual feature $v_{i}^{\mathrm{cls}}$.
	
		\subsubsection{Text Encoder} \label{subsec322}
	
			We adopt a casual masked transformer to model text inputs. Texts are tokenized into an embedding sequence of length $S+1$ represented as $W_{\mathrm{emb},i}$, and attain encoded features:
			\[
			W_{i} = [w_{i}^{1}, \ldots, w_{i}^{S}, w_{i}^{\mathrm{cls}}] \in \mathbb{R}^{(S+1) \times C_{2}}
			\]
			with $w^{\mathrm{cls}}$ being the CLS text feature and $C_{2}$ denotes the text embedding dimension.

		\subsubsection{Text-guided Image Decoder} \label{subsec323}
	
	
			To recover masked patches from $\widetilde{P}_{\mathrm{emb},i}^{(\mathrm{l},n)}$ and make reconstruction aware of textual semantics, we adopt a text-guided imaged decoder. This process ensures that the image reconstruction is \textit{context-aware}, refining the visual features based on verbal descriptions. Specifically, we project $V_{i,n}^{(\mathrm{l})}$ to the text embedding space via $\Pi_{v\to t}:\mathbb{R}^{C_1}\to\mathbb{R}^{C_2}$, and introduce $mP$ learnable mask tokens $P_{\text{mask}}\in\mathbb{R}^{mP\times C_2}$. The decoder input for local crop $n$ is:
			\[
			D_{i,n}^{\mathrm{in}}=[P_{\text{mask}};\ \Pi_{v\to t}(V_{i,n}^{(\text{l})})]\in\mathbb{R}^{(P+1)\times C_2}
			\]
			
			To stabilize fuse textual context, a gate-based fusion mechanism is used: 
			\[
			U_{i,n}^{\mathrm{img}} = \mathrm{Gate_{img}}(W_{i}, D_{i,n}^{\mathrm{in}}) \in \mathbb{R}^{(P+1) \times C_{2}}
			\]
			
			As utilizing $D_{i,n}^{\mathrm{in}}$ as the query and $U_{i,n}^{\mathrm{img}}$ as the key and value, we leverage textual semantic information to compel the masked image reconstruction process to perceive verbal cues, and attained decode:
			\[
			\{\widetilde{X}_{i,n}^{p}\}_{p \in \overline{M}_{i,n}},\ z_{i,n} = \mathrm{ImgDec}(D_{i,n}^{\text{in}},\ U_{i,n}^{\text{img}}),
			\]
			where $\widetilde{X}_{i,n}^p \in \mathbb{R}^{(P+1)\times C_2}$ is the predicted pixel value for the $p$-th masked patch, and $z_{i,n}\in\mathbb{R}^{C_2}$ is a sub-image-level representation used for fine-grained alignment and revision. Through this \textit{text-guided reconstruction}, the image features are refined to be semantically consistent with the accompanying text.
	
		\subsubsection{Image-guided Text Decoder} \label{subsec324}
	
			Symmetrically, we design an image-guided text decoder aims to refine the text representation by incorporating multi-scale non-verbal visual cues. It allows the text semantics to be grounded in and disambiguated by the rich context provided by the image, addressing the inherent ambiguity of text-only analysis.
			
			We concatenate multi-scale visual features from the global image and all local sub-images:
			\[
			V_i^{\text{all}}=[\Pi_{v\to t}(V_i^{\text{(g)}}),\ \Pi_{v\to t}(V_{i,1}^{(\text{l})}),\dots,\Pi_{v\to t}(V_{i,n}^{(\text{l})})]\in\mathbb{R}^{nP\times C_2}
			\]
			as the key and value, and use non-CLS text tokens $W_{i}^{1:S}$ as the query. A cross-attention mechanism then fuses these non-verbal visual cues into the text representation:
			\[
			\widetilde{W}_i = \mathrm{TextDec}(W_{i}^{1:S},\ V_i^{\text{all}})
			\]
			
			The output $\widetilde{W}_{i} \in \mathbb{R}^{S \times C_2}$ is a visually-refined text feature, which is then passed through a lightweight MLP for the final prediction $\widehat{y}_{i}$. This step constitutes the \textit{image-guided refinement} of the text modality.

	\subsection{Loss Functions} \label{subsec33}


		\subsubsection{Reconstruction Loss} \label{subsec331}
		
			The reconstruction loss supervises the text-guided image decoder to recover masked patches in high-resolution sub-images. By minimizing the \textit{mean squared error} between the the masked image tokens and the reconstructed tokens, this loss enhances the model's capability to capture fine-grained visual details:
			\begin{equation}
				\mathcal{L}_{{rec}} = \frac{1}{N} \sum_{i=1}^{N} \left( \frac{1}{\sum_{n} |\overline{M}_{i,n}|} \sum_{n} \sum_{p \in \overline{M}_{i,n}} \| X_{i,n}^{p} - \widetilde{X}_{i,n}^{p} \|_{2} \right)
			\end{equation}
			where $X_{i,n}^{p}$ and $\widetilde{X}_{i,n}^{p}$ are the RGB pixel value and predicted pixel values for the $p$-th masked patch in the $n$-th sub-image, $|\overline{M}_{i,n}|$ denotes the number of masked patches.

		\subsubsection{Image-Text Contrastive Loss} \label{subsec332}
	
			To align the global representations of image and text modalities in a shared semantic space, we employ an image-text contrastive loss. This loss encourages matched image-text pairs to have similar representations while pushing unmatched pairs apart:
			\begin{equation}
				\begin{aligned}
					\mathcal{L}_{{itc}} = \frac{1}{2N} \Bigg[ & \sum_{i=1}^{N} -\log \left( \frac{\exp \left( \langle v, w \rangle / \tau 	\right)}{\sum_{j=1}^{N} \exp \left( \langle v, w_{j} \rangle / \tau \right)} \right) \Bigg] \\
					+ \frac{1}{2N} \Bigg[ & \sum_{i=1}^{N} -\log \left( \frac{\exp \left( \langle w, v \rangle / \tau 	\right)}{\sum_{j=1}^{N} \exp \left( \langle w, v_{j} \rangle / \tau \right)} \right) \Bigg]
				\end{aligned}
			\end{equation}
			where $v = v_{i}^{{cls}}$ and $w = w_{i}^{\mathrm{cls}}$ are the normalized global features from the low-resolution image and text, respectively, $\langle \cdot, \cdot \rangle$ denotes the inner product, and $\tau$ is a temperature parameter.

		\subsubsection{Local-Global Semantic Similarity Loss} \label{subsec333}

			It may cause the reconstructed features to deviate from global visual semantics, although MIM focuses on local details. We introduce a cross-scale consistency loss to maintain alignment between local reconstructed features and global visual representations, resulting in over rely local information. To enforce consistency between the reconstructed local features and the global visual representations, we perform learnable weighted aggregation on the sub-image-level global representation $\{z_{i,n}\}_{n=1}^{4}$ to obtain:
			\[
			\begin{aligned}
				e_{i,n} &= u^{T} \tanh \left( z_{i,n} W_{z} + b \right)\\
				\alpha_{i,n} &= \left[ \mathrm{softmax} \left( [e_{i,1}, \ldots, e_{i,4}] \right) \right]_{n} \\
				P_{\mathrm{agg},i} &= \mathrm{MLP} \left( \sum_{n=1}^{4} \alpha_{i,n} z_{i,n} \right)
			\end{aligned}
			\]
			here, $W_{z}$, $u$, $b$ are learnable parameters, and the $\mathrm{MLP}$ projects the aggregated feature into same semantic space as $v_{i}^{\mathrm{cls}}$. After normalizing to all features, we compute:
			\begin{equation}
				\mathcal{L}_{{si}} = \frac{1}{N} \sum_{i=1}^{N} \| v_{i}^{\mathrm{cls}} - P_{\mathrm{agg},i} \|_{2}^{2}
			\end{equation}
			This loss ensures that the reconstructed local pixel features from high-resolution sub-images remain consistent with the global image semantics from low-resolution images, achieving a balance between local and global features.

		\subsubsection{Cross-Modal Feature-Distribution Consistency Loss} \label{subsec334}

			The pixel-level reconstruction ($\mathcal{L}_{{rec}}$) and semantic-level alignment ($\mathcal{L}_{{itc}}$) objectives may conflict. To harmonize them, we enforce distribution consistency between the text-to-reconstructed-image and text-to-global-image similarity distributions:
			\begin{equation}
				\mathcal{L}_{{dc}} = \frac{1}{N} \sum_{i=1}^{N} \left[ \text{KL}(S_{{t2l}} \| S_{{t2g}}) + \lambda \cdot \text{H}(S_{{t2l}}) \right]
			\end{equation}
			where $S_{{t2l}} = S(P_{\mathrm{agg},i}, w_{i}^{\mathrm{cls}})$ and $S_{{t2g}} = S(v_{i}^{\mathrm{cls}}, w_{i}^{\mathrm{cls}})$ are similarity distributions, $\mathrm{KL}$ denotes the relative entropy loss function, $\mathrm{H}$ represents the entropy regularization term for robustness, and $\lambda$ is a weighting factor.

		\subsubsection{Classification Loss} \label{subsec335}

			For downstream tasks, such as desire understanding, emotion recognition, and sentiment analysis, we use the standard cross-entropy loss:
			\begin{equation}
				\mathcal{L}_{{cls}} = \frac{1}{N} \sum_{i=1}^{N} \mathrm{CrossEntropy}(y_{i}, \widehat{y}_{i})
			\end{equation}
			where $y$ is the ground-truth label and $\widehat{y}_{i}$ is the predicted label.	

	\subsection{Two-stage Training Strategy} \label{subsec34}

		To balance semantic alignment with pixel-level reconstruction, we adopt a two-stage training strategy and selectively freeze or unfreeze model components in each stage to steer learning.

		\subsubsection{Pre-training Stage} \label{subsec341}
		
			The pre-training stage focuses on masked image modeling, giving priority to training the image encoder and the text-guided image decoder. In this stage, we freeze the image-guided text decoder and the MLP, and only update the text encoder, the image encoder, and the text-guided image decoder. This design allows us to extract fine-grained details from high-resolution sub-images while constraining modal consistency at the semantic level. As discussed in Section~\ref{subsec333} and Section~\ref{subsec334}, we also consider the sub-image-level global visual representation of reconstructed image tokens and their cross-modal alignment. Therefore, the overall loss used in this stage is:
			\begin{equation}
				{{\mathcal{L}}_{p}}={{\lambda }_{rec}}{{\mathcal{L}}_{rec}}+{{\lambda }_{si}}{{\mathcal{L}}_{si}}+{{\lambda }_{dc}}{{\mathcal{L}}_{dc}}+{{\lambda }_{itc}}{{\mathcal{L}}_{itc}}
			\end{equation}
			where ${{\lambda}_{rec}}$, ${{\lambda}_{si}}$, ${{\lambda}_{dc}}$ and ${{\lambda}_{itc}}$ are hyperparameter weights that balance the contributions of each loss term.

		\subsubsection{Fine-tuning Stage} \label{subsec342}

			The fine-tuning stage targets downstream tasks such as desire understanding. Its goal is to train the image-guided text decoder to fully leverage both local and global features produced by the image encoder, and to use an MLP to map the fused multimodal representation to task-specific outputs. During fine-tuning we freeze the image encoder and text-guided image decoder, and train the text encoder, the image-guided text decoder, and the MLP. The overall loss for this stage is:
			\begin{equation}
				{{\mathcal{L}}_{f}}={{\lambda }_{cls}}{{\mathcal{L}}_{cls}}+{{\lambda }_{itc}}{{\mathcal{L}}_{itc}}
			\end{equation}
			where ${{\lambda }_{cls}}$ and ${{\lambda }_{itc}}$ are hyperparameter weights. Retaining the $\text{ITC}$ term helps preserve cross-modal alignment stability during fine-tuning.	

			\begin{table}[bp]
				\centering
				\caption{Statistics of MSED Dataset}
				\label{tab:msed_stats_summary}
				\resizebox{0.8\columnwidth}{!}{%
					\begin{tabular}{c c c c}
						\toprule
						\textbf{Train} & \textbf{Validation} & \textbf{Test} & \textbf{Total} \\
						\midrule
						6,127 (66.7\%) & 1,021 (11.1\%) & 2,042 (22.2\%) & 9,190 \\
						\bottomrule
					\end{tabular}%
				}
			\end{table}

\section{Experiments}
\label{sec4}

	\subsection{Experiment Setup}
	\label{subsec41}

		\subsubsection{Dataset} \label{subsec411}
		
			\begin{table*}[tbp!]
				\centering
				\caption{Comparison of SyDES and other SOTA methods on the MSED dataset}
				\label{tab:sota_comparison}
				\resizebox{1.00\textwidth}{!}{%
					\begin{NiceTabular}{c | *{4}{c} *{4}{c} *{4}{c}}
						\toprule
						\multirow{2}{*}{\textbf{Method}} & 
						\multicolumn{4}{c}{\textbf{Desire Understanding}} & 
						\multicolumn{4}{c}{\textbf{Emotion Recognition}} & 
						\multicolumn{4}{c}{\textbf{Sentiment Analysis}} \\
						\cmidrule(lr){2-5} \cmidrule(lr){6-9} \cmidrule(l){10-13}
						& P & R & F1 & Acc & P & R & F1 & Acc & P & R & F1 & Acc \\
						\midrule
						DCNN+AlexNet \cite{Jia2022BeyondEA} & 59.42 & 52.02 & 52.35 & - & 49.56 & 42.77 & 43.76 & - & 71.02 & 70.09 & 70.31 & - \\
						DCNN+ResNet \cite{Jia2022BeyondEA} & 56.34 & 50.64 & 52.89 & - & 62.93 & 59.12 & 60.48 & - & 74.73 & 74.73 & 74.64 & - \\
						BiLSTM+AlexNet \cite{Jia2022BeyondEA} & 67.80 & 68.00 & 67.67 & - & 71.17 & 70.70 & 70.89 & - & 78.73 & 79.22 & 78.89 & - \\
						BERT+AlexNet \cite{Jia2022BeyondEA} & 80.84 & 75.50 & 77.17 & - & 78.06 & 78.19 & 78.10 & - & 83.22 & 83.11 & 83.16 & - \\
						Multimodal Transformer \cite{Jia2022BeyondEA} & 81.92 & 80.20 & 80.92 & - & 81.62 & 81.61 & 81.53 & - & 83.56 & 83.45 & 83.50 & - \\
						M3GAT \cite{Zhang2023M3GATAM} & - & - & - & - & 82.53 & 81.51 & 81.97 & - & 84.66 & 85.15 & 84.85 & - \\
						MMTF-DES \cite{Aziz2025MMTFDESAF} & \textbf{84.23} & \underline{82.01} & \underline{83.11} & \underline{86.97} & \underline{84.39} & \underline{84.64} & \underline{84.26} & \underline{84.13} & \underline{88.27} & \underline{88.68} & \underline{88.44} & \underline{88.44} \\
						SyDES (Ours) & \underline{84.09} & \textbf{84.07} & \textbf{84.02} & \textbf{88.32} & \textbf{84.92} & \textbf{84.81} & \textbf{84.74} & \textbf{85.96} & \textbf{89.28} & \textbf{89.13} & \textbf{89.19} & \textbf{89.37} \\
						\hline
						\%Gains & \textcolor{red!50}{-0.20} & \textcolor{blue!50}{+2.50} & \textcolor{blue!50}{+1.10} & \textcolor{blue!50}{+1.60} & \textcolor{blue!50}{+0.60} & \textcolor{blue!50}{+0.20} & \textcolor{blue!50}{+0.60} & \textcolor{blue!50}{+2.20} & \textcolor{blue!50}{+1.10} & \textcolor{blue!50}{+0.50} & \textcolor{blue!50}{+0.90} & \textcolor{blue!50}{+1.10} \\
						\bottomrule
					\end{NiceTabular}%
				} %
			\end{table*}
			
			\begin{table}[tbp!]
				\centering
				\caption{Comparison between SyDES and different modality baseline models on the MSED dataset}
				\label{tab:method_comparison}
				\renewcommand{\arraystretch}{0.80}
				\resizebox{\columnwidth}{!}{%
					\begin{NiceTabular}{c | c | c | c c c}
						\toprule
						\textbf{Task} & \textbf{Method} & \textbf{Modality} & \textbf{P} & \textbf{R} & \textbf{F1} \\
						\midrule
						\multirow{4}{*}{Desire Understanding} 
						& BERTweet & Text & 77.11 & 81.19 & 78.86 \\
						& ResNet & Image & 49.97 & 49.35 & 49.20 \\
						& SyDES-B & Multimodal & 80.77 & 67.43 & 72.27 \\
						& SyDES & Multimodal & \textbf{84.09} & \textbf{84.07} & \textbf{84.02} \\
						\midrule
						\multirow{4}{*}{Emotion Recognition} 
						& BERTweet & Text & 80.99 & 77.15 & 78.34 \\
						& ResNet & Image & 58.74 & 54.67 & 56.40 \\
						& SyDES-B & Multimodal & 81.04 & 80.66 & 80.80 \\
						& SyDES & Multimodal & \textbf{84.92} & \textbf{84.81} & \textbf{84.74} \\
						\midrule
						\multirow{4}{*}{Sentiment Analysis} 
						& BERTweet & Text & 82.25 & 83.62 & 82.49 \\
						& ResNet & Image & 70.85 & 70.61 & 70.64 \\
						& SyDES-B & Multimodal & 89.09 & 85.58 & 86.50 \\
						& SyDES & Multimodal & \textbf{89.28} & \textbf{89.13} & \textbf{89.19} \\
						\bottomrule
					\end{NiceTabular}%
				}
			\end{table}

			To validate our method, we use the MSED dataset \cite{Jia2022BeyondEA}, the first multimodal multi-task benchmark for sentiment analysis, emotion recognition, and desire understanding. MSED comprises $9{,}190$ English image–text pairs collected from social media (e.g., Twitter, Getty Images, Flickr). Each sample is annotated with a sentiment label (\textit{positive, neutral, negative}), an emotion label (\textit{happiness, sad, neutral, disgust, anger, and fear}), and a desire label (\textit{family, romance, vengeance, curiosity, tranquility, social-contact, and none}). The data are split into train, validation, and test set. Detailed statistics are given in Table~\ref{tab:msed_stats_summary} and the detailed statistics are provided in Appendix~\ref{app:msed_stats}.

		\subsubsection{Evaluation Metrics} \label{subsec412}
		
			All three downstream tasks are classification problems. We therefore report standard classification metrics: Precision ($\text{P}$), Recall ($\text{R}$), Macro-F1-score ($\text{F1}$), and Weighted Accuracy ($\text{Acc}$).

		\subsubsection{Training Details} \label{subsec413}
		
			All experiments run on NVIDIA V100 with CUDA $11.0$ and PyTorch $2.1.2$ \cite{Paszke2019PyTorchAI}. We initialize the image encoder and the text encoder from CLIP \cite{Radford2021LearningTV} pre-training weights provided by OpenAI\footnote{\url{https://github.com/openai/CLIP}}, and initialize the text-guided image decoder from pre-training weights\footnote{\url{https://huggingface.co/laion/mscoco_finetuned_CoCa-ViT-L-14-laion2B-s13B-b90k}} in \cite{ilharco_gabriel_2021_5143773}. For complete training hyper-parameter settings are provided in Appendix~\ref{app:training}.

		\subsubsection{Baseline} \label{subsec414}

			To facilitate subsequent ablation analysis and comparison, we use the SyDES architecture trained directly with classification loss without pre-training stage. We denote this variant as \textbf{SyDES-B} (the SyDES baseline).

		\subsection{Comparison with different modality baseline models} \label{subsec42}
		
			We compared different modality baseline models on three downstream tasks, including desire understanding, emotion recognition, and sentiment analysis, to evaluate the effectiveness of our proposed SyDES. For the textual modality, we employed \textbf{BERTweet} \cite{Nguyen2020BERTweetAP} model as the baseline modal since text data are annotated from social media platforms. For the image modality, we used the classic backbone network \textbf{ResNet} \cite{He2015DeepRL}. The multimodal baseline model, SyDES-B, was included to demonstrate the advantage of multimodal fusion. We also present the results of our proposed SyDES method.
			
			The experiment results for different modality baseline models are shown in Table~\ref{tab:method_comparison}. Analysis of these results leads to three main conclusions: (1) Multimodal models consistently outperform unimodal models across mostly tasks. For instance, in emotion recognition, SyDES-B improved the F1-score by 3.14\% gains over BERTweet, indicating the benefit of leveraging multiple modalities in sentiment-related tasks. (2) The unimodal image model (e.g., ResNet) consistently underperformed compared to the unimodal text model (e.g., BERTweet), suggesting limitations in capturing fine-grained visual semantics and underutilization of non-verbal cues. (3) Our proposed SyDES consistently surpassed SyDES-B in all tasks, validating the effectiveness of the non-verbal cues mining mechanism introduced during pre-training.

	\subsection{Comparison with state-of-the-art methods} \label{subsec43}

		\begin{table*}[tbp]
			\centering
			\caption{Ablation study on the loss functions used in the two training stages}
			\label{tab:loss_ablation}
			\resizebox{1.00\textwidth}{!}{%
				\begin{NiceTabular}{cccc | cc | *{4}{c} *{4}{c} *{4}{c}}
					\toprule
					\multicolumn{4}{c}{\textbf{Pre-training}} & 
					\multicolumn{2}{c}{\textbf{Fine-tuning}} & 
					\multicolumn{4}{c}{\textbf{Desire Understanding}} & 
					\multicolumn{4}{c}{\textbf{Emotion Recognition}} & 
					\multicolumn{4}{c}{\textbf{Sentiment Analysis}} \\
					\cmidrule(lr){1-4} \cmidrule(lr){5-6} \cmidrule(lr){7-10} \cmidrule(lr){11-14} \cmidrule(l){15-18}
					$\mathcal{L}_{\text{rec}}$ & $\mathcal{L}_{{itc}}$ & $\mathcal{L}_{{si}}$ & $\mathcal{L}_{{dc}}$ & $\mathcal{L}_{{itc}}$ & $\mathcal{L}_{{cls}}$ & P & R & F1 & Acc & P & R & F1 & Acc & P & R & F1 & Acc \\
					\midrule
					$\times$ & $\times$ & $\times$ & $\times$ & $\times$ & $\checkmark$ & 80.77 & 67.43 & 72.27 & 80.51 & 81.04 & 80.66 & 80.80 & 82.61 & 89.09 & 85.58 & 86.50 & 87.12 \\
					$\checkmark$ & $\times$ & $\times$ & $\times$ & $\checkmark$ & $\checkmark$ & 76.17 & 72.56 & 74.12 & 81.15 & 76.78 & 75.13 & 75.82 & 77.62 & 80.57 & 81.19 & 80.79 & 80.80 \\
					$\checkmark$ & $\checkmark$ & $\times$ & $\times$ & $\checkmark$ & $\checkmark$ & 81.43 & 81.77 & 81.44 & 86.19 & \underline{84.85} & \underline{84.95} & \underline{84.66} & \underline{85.21} & \underline{89.26} & \underline{88.80} & \underline{88.99} & \underline{89.18} \\
					$\checkmark$ & $\checkmark$ & $\checkmark$ & $\times$ & $\checkmark$ & $\checkmark$ & \underline{82.28} & \underline{82.33} & \underline{82.24} & \underline{86.78} & 84.17 & 81.91 & 82.88 & 84.53 & 87.45 & 87.66 & 87.55 & 87.71 \\
					$\checkmark$ & $\checkmark$ & $\checkmark$ & $\checkmark$ & $\times$ & $\checkmark$ & 78.55 & 75.89 & 77.09 & 83.74 & \underline{84.85} & 81.75 & 83.21 & 84.43 & 86.46 & 85.93 & 86.16 & 86.44 \\
					\rowcolor{lightgray!30}
					$\checkmark$ & $\checkmark$ & $\checkmark$ & $\checkmark$ & $\checkmark$ & $\checkmark$ & \textbf{84.09} & \textbf{84.07} & \textbf{84.02} & \textbf{88.32} & \textbf{84.92} & \underline{84.81} & \textbf{84.74} & \textbf{85.96} & \textbf{89.28} & \textbf{89.13} & \textbf{89.19} & \textbf{89.37} \\
					\bottomrule
				\end{NiceTabular}%
			}
		\end{table*}
		
		The comparative performance of our proposed SyDES on test data against other SOTA methods across all tasks is presented in Table~\ref{tab:sota_comparison}. The results indicate that SyDES achieves competitive performance across all three tasks. In terms of the primary metric F1-score, our proposed SyDES surpassed the previous best model, MMTF-DES \cite{Aziz2025MMTFDESAF}, by 1.1\%, 0.6\%, and 0.9\% gains, respectively. It is worth noting that MMTF-DES relies on integrating multiple multimodal Transformer encoders (e.g., ViLT and VAuLT), which entails considerably higher training costs. In contrast, SyDES extracts non-verbal cues from images effectively while maintaining lower computational overhead, yielding pronounced gains on desire understanding, which depends more heavily on non-verbal cues.
		
		We further observe that sentiment analysis generally yields higher performance among the three tasks. A possible reason for this difference in performance may be attributed to the nature of the three tasks. Sentiment analysis aim to identify the overall emotion or opinion expressed in an image-text pair. In contrast, desire understanding and emotion recognition require fine-grained detection of specific signals such as a person’s gestures or facial expressions that are inherently embedded in images. For example, there exists the exaggerated motion and frightened expression of the man in Figure~\ref{fig:dataset} (b). These subtle cues are inherently more challenging to capture. Therefore, sentiment analysis may be an easier and more straightforward task for model, while desire understanding and emotion recognition may be more complicated and nuanced. Our proposed SyDES enhances local detail extraction, resulting in particularly notable gains in desire understanding, but its performance remains slightly below that of sentiment analysis. This suggests there is still room for improvement and underscores the need for further research into desire understanding and emotion recognition.

		\begin{table}[tbp]
			\centering
			\caption{Ablation study of proposed components.}
			\label{tab:abl_module}
			\resizebox{\columnwidth}{!}{%
				\begin{NiceTabular}{l | l | cccccc}
					\toprule
					\multirow{2}{*}{\textbf{Step}} & \multirow{2}{*}{\textbf{Configuration}} & \multicolumn{2}{c}{\textbf{Desire}} & \multicolumn{2}{c}{\textbf{Emotion}} & \multicolumn{2}{c}{\textbf{Sentiment}} \\
					\cmidrule(lr){3-4} \cmidrule(lr){5-6} \cmidrule(l){7-8}
					& & F1 & Acc & F1 & Acc & F1 & Acc \\
					\midrule
					0 & \textit{Baseline (SyDES-B)} & 72.27 & 80.51 & 80.80 & 82.61 & 86.50 & 87.12 \\
					1 & + Mixed-scale image strategy & 75.49 & 79.97 & 78.70 & 81.49 & 86.37 & 86.24 \\
					2 & + Reconstruction decoder & 81.51 & 84.96 & 84.40 & 85.26 & 89.03 & 89.08 \\
					\rowcolor{lightgray!30} 3 & + Aggregator & \textbf{84.02} & \textbf{88.32} & \textbf{84.74} & \textbf{85.96} & \textbf{89.19} & \textbf{89.37} \\
					\bottomrule
				\end{NiceTabular}%
			}
		\end{table}

	\subsection{Ablation Studies} \label{subsec44}
	
		\subsubsection{Loss Functions} \label{subsec441}

			We adopt a two-stage training strategy to facilitate cross-modal fusion between textual features and both local and global visual features. As summarized in Table~\ref{tab:loss_ablation}, we systematically evaluate the impact of different loss combinations on the three sub-tasks. The first row corresponds to the model fin-tuned directly without pre-training stage (i.e., SyDES-B as described in Section~\ref{subsec414}). During pre-training, four loss functions are used, including ${{\mathcal{L}}_{rec}}$, ${{\mathcal{L}}_{si}}$, ${{\mathcal{L}}_{dc}}$, and ${{\mathcal{L}}_{itc}}$. Results show that using only ${{\mathcal{L}}_{rec}}$ for masked image modeling leads to performance even worse than SyDES-B. This indicates that reconstructing high-resolution sub-images alone may introduce a semantic mismatch with the low-resolution image due to inadequate modal alignment, resulting in modal inconsistency that harms fine-tuning. Gradually incorporating ${{\mathcal{L}}_{itc}}$, ${{\mathcal{L}}_{si}}$, and ${{\mathcal{L}}_{dc}}$ consistently improves performance across all tasks. For example, in desire understanding, the F1-score increases to 81.44\% with ${{\mathcal{L}}_{itc}}$, to 82.24\% with ${{\mathcal{L}}_{si}}$, and finally to 84.02\% with ${{\mathcal{L}}_{dc}}$. This suggests that cross-modal alignment and semantic/consistency constraints are critical to bridging the gap between reconstructed sub-images and the global image semantics. More detailed result can be found in Appendix~\ref{app:compelete_diff_losses}.
			
			During fine-tuning, two loss functions are used, including ${{\mathcal{L}}_{itc}}$ and ${{\mathcal{L}}_{cls}}$. We observe that that ${{\mathcal{L}}_{itc}}$ is important to preserve pre-training gains. Removing ${{\mathcal{L}}_{itc}}$ during fine-tuning causes the F1-scores to drop from 84.02\%, 84.74\%, and 89.19\% to 77.09\%, 83.21\%, and 86.16\%, respectively. In conclusion, using all proposed loss functions yields the best performance across desire understanding, emotion recognition, and sentiment analysis, confirming the complementary effects of the loss terms.
		
		\subsubsection{Contribution of proposed modules} \label{subsec442}
		
			As shown in Table~\ref{tab:abl_module}, the results demonstrate that the complete model achieves the best performance across three tasks, with the F1 score for desire understanding significantly increasing from 72.27\% to 84.0\%. It is worth noting that mixed-scale image strategy alone leads to a performance drop in desire understanding, indicating that simple modality dose not fully exploit fine-gained semantic alignment. Introducing the reconstruction decoder substantially improves desire understanding, demonstrating that fine-grained reconstruction effectively enhances the model's sensitivity to local semantics. Moreover, incorporating the aggregator along with consistency losses further improves all metrics, highlighting the importance of global-local semantic coordination for effective learning.
			
		\subsubsection{Ratio of Masked Tokens} \label{subsec443}
		
			As shown in Table~\ref{tab:mask_ratio_ablation}, we investigate the impact by setting different ratios of masked tokens. Specifically, 0.25, 0.50, 0.75, and 0.90 are tested. Experimental results demonstrate that a masking ratio of 75\% yields the best average performance across most tasks. Although a ratio of 25\% achieves slightly better results in emotion recognition, the 75\% ratio is overall more suitable considering its substantially lower computational cost while maintain competitive performance.
			
			\begin{table*}[tbp]
				\centering
				\caption{Ablation study on the ratio of masked tokens}
				\label{tab:mask_ratio_ablation}
				\resizebox{1.00\textwidth}{!}{%
					\begin{NiceTabular}{c | *{4}{c} *{4}{c} *{4}{c}}
						\toprule
						\multirow{2}{*}{\textbf{Ratio of Masked Tokens}} & 
						\multicolumn{4}{c}{\textbf{Desire Understanding}} & 
						\multicolumn{4}{c}{\textbf{Emotion Recognition}} & 
						\multicolumn{4}{c}{\textbf{Sentiment Analysis}} \\
						\cmidrule(lr){2-5} \cmidrule(lr){6-9} \cmidrule(l){10-13}
						& {P} & {R} & {F1} & {Acc} & {P} & {R} & {F1} & {Acc} & {P} & {R} & {F1} & {Acc} \\
						\midrule
						0.25 & \underline{83.13} & 81.80 & \underline{82.23} & \underline{86.82} & \textbf{85.94} & \textbf{85.30} & \textbf{85.55} & \underline{85.94} & \underline{88.90} & \underline{89.05} & \underline{88.87} & \underline{88.83} \\
						0.50 & 81.87 & 81.69 & 81.69 & 86.48 & 84.36 & 84.35 & 84.19 & 85.06 & 87.64 & 88.18 & 87.88 & 88.06 \\
						\rowcolor{lightgray!30}
						0.75 & \textbf{84.09} & \textbf{84.07} & \textbf{84.02} & \textbf{88.32} & \underline{84.92} & \underline{84.81} & \underline{84.74} & \textbf{85.96} & \textbf{89.28} & \textbf{89.13} & \textbf{89.19} & \textbf{89.37} \\
						0.90 & 81.31 & \underline{82.82} & 82.02 & 86.48 & 84.32 & 83.31 & 83.78 & 85.41 & 88.54 & 87.93 & 88.17 & 88.39 \\
						\bottomrule
					\end{NiceTabular}%
				}
			\end{table*}

	\subsection{Visualization} \label{subsec45}
	
		\subsubsection{Cross-Modal Attention Heatmap Analysis} \label{subsec451}
	
			\begin{figure}[tp]
				\centering
				\includegraphics[width=\columnwidth]{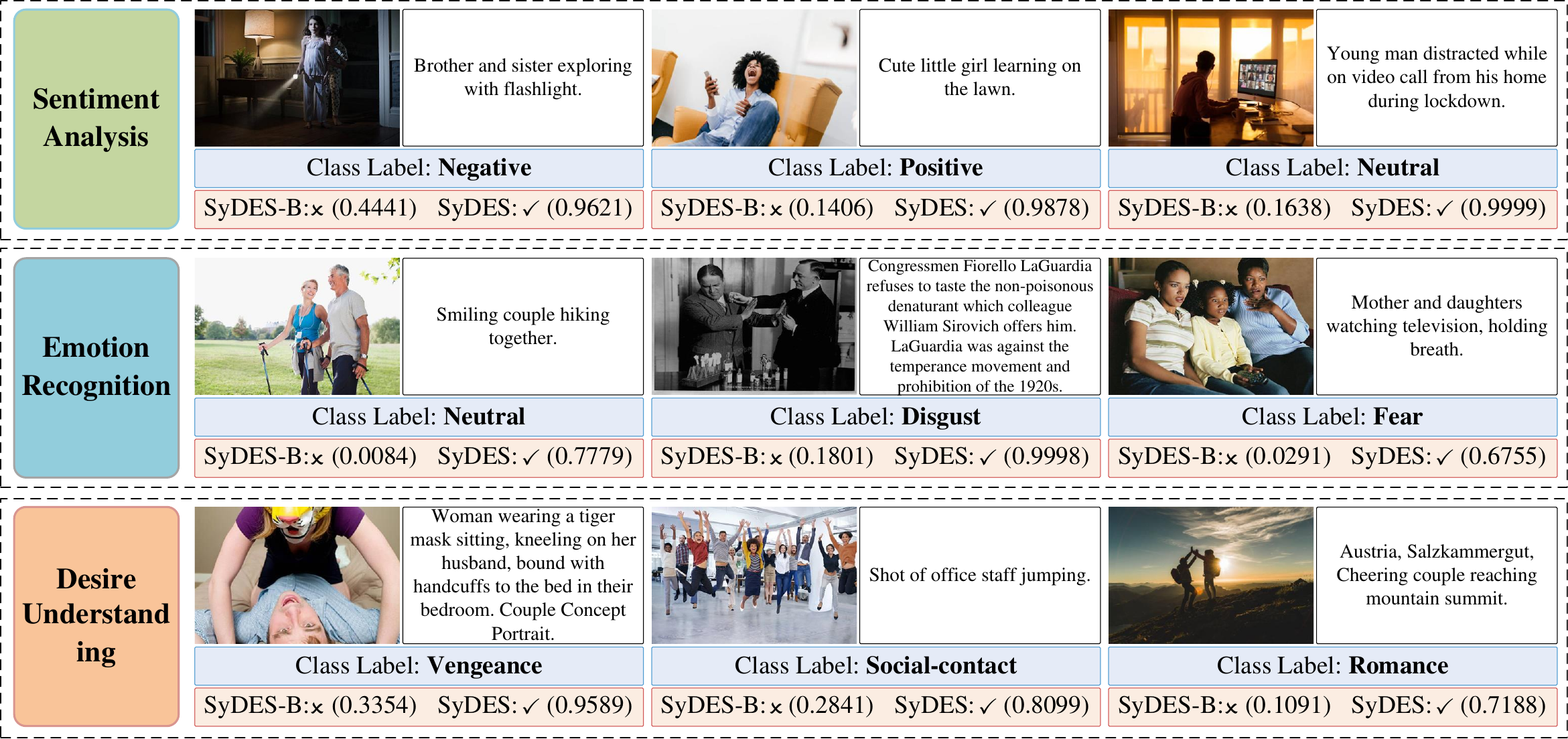}
				\caption{Performance analysis of our proposed SyDES on desire understanding, emotion recognition, and sentiment analysis tasks. $\times$ indicates correct classification and $\checkmark$ represents misclassification.}
				\label{fig:example}
			\end{figure}
		
\begin{figure}[tp]
				\centering
				\includegraphics[width=\columnwidth]{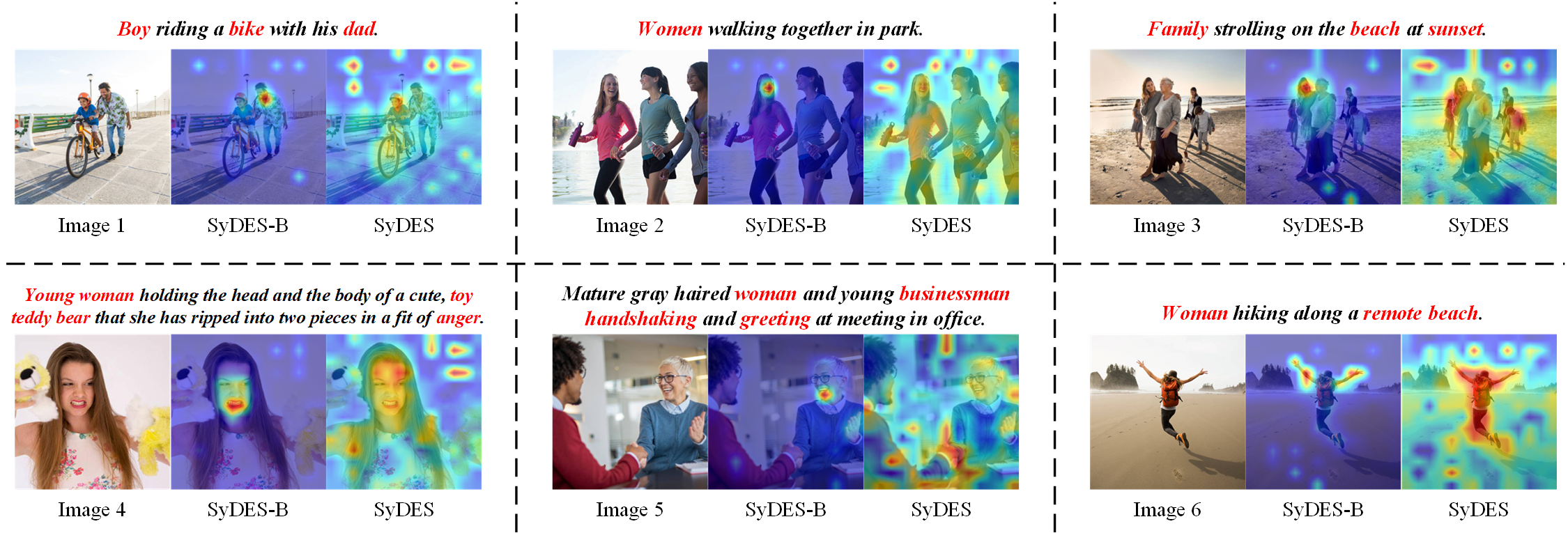}
				\caption{Qualitative analysis of attention localization map in our proposed SyDES. We visualize the attention localization map from the last layer in the image encoder, reflecting the model’s cross-modal perception of text.}
				\label{fig:attns}
			\end{figure}

			To intuitively illustrate the perception ability of our proposed SyDES toward textual and image modalities, we visualized the attention localization map for the last layer in the image encoder. As depicted in Figure~\ref{fig:attns}, our SyDES focuses more accurately on fine-grained visual regions corresponding to informative words in the text. For example, in the case of Image 1, SyDES shows concentrated attention on regions related to “\textit{\textbf{boy}}” and “\textit{\textbf{bike}}”, whereas SyDES-B neglects these essential details. Similarity, in the case of Image 3, SyDES-B perceives the concept of “\textit{\textbf{family}}” vaguely, while our proposed SyDES clearly identifies the five-person “\textit{\textbf{family}}” and the “\textit{\textbf{beach}}” scene. 

		\subsubsection{Reconstructed Image Visualization} \label{subsec452}
		
\begin{figure}[tp]
				\centering
				\includegraphics[width=\columnwidth]{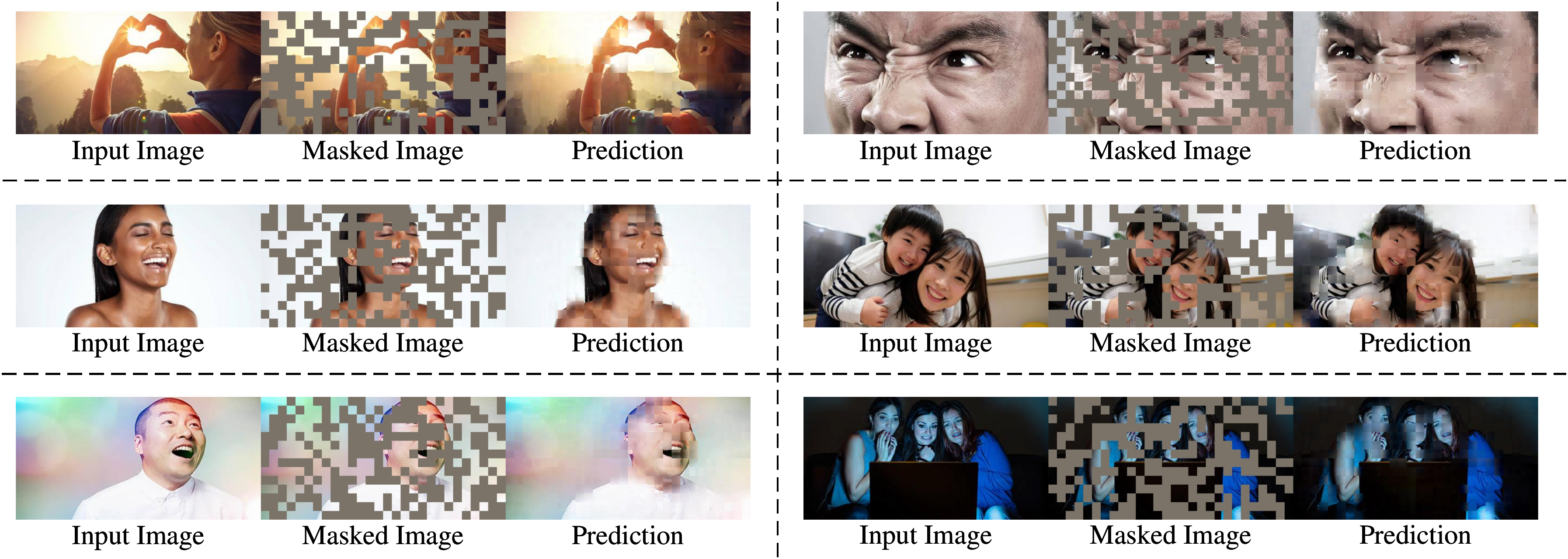}
				\caption{Visualization of reconstructed images using our proposed SyDES.}
				\label{fig:masked}
			\end{figure}

			We reconstructed and stitched the high-resolution sub-images, and visualized the masked image patches to inspect the practical effect in Figure~\ref{fig:masked}. Each sample consists of the original image, the randomly masked image, and the reconstructed image by the text-guided image decoder. The results show that, even with complex image content (e.g., \textit{\textit{people}, \textit{gesture}, \textit{lighting}, and \textit{natural scenes}} in the example 1), our proposed SyDES can reconstruct contextual content reasonably well, owing to masked image modeling using high-resolution sub-images and text-guided decoding. 
			
		\subsubsection{Performance Analysis} \label{subsec453}
			
			To better understand the advantage of our proposed SyDES, we selected representative examples from the MSED dataset across three sub-tasks.  As illustrated in Figure~\ref{fig:example}, we compare the predictions of SyDES-B and our proposed SyDES, including example stimuli, ground-truth labels, and predicted probabilities. Across all three tasks, our proposed SyDES produces correct predictions with high confidence. For instance, in Example 1 of the sentiment analysis task, the image caption is “\textit{\textbf{Brother and sister exploring with flashlight.}}”, it provides limited information. But our proposed SyDES achieved a high confidence score of 96.21\% by effectively leveraging visual cues. 
		
\section{Conclusion}
\label{sec6}

	To address the under-exploration of desire understanding and the inadequate use of non-verbal cues, we propose a symmetric bidirectional multimodal learning framework for desire, emotion, and sentiment recognition. The framework employs a mixed-scale image strategy, using masked image modeling to boosting extract fine-grained local features from high-resolution sub-images and maintaining global context from low-resolution images. Furthermore, its symmetrical cross-modal decoders, including a text-guided image decoder and an image-guided text decoder, enable bidirectional interaction that refines both visual and textual representations. To harmonize these objectives, a set of dedicated loss functions  are used, facilitating consistent learning across modalities. Extensive experiments on the MSED dataset demonstrate the effectiveness of leveraging non-verbal cues.

\begin{acks}
	This work is supported by the National Natural Science Foundation of China (Grant No. 62272188). 
\end{acks}

\bibliographystyle{ACM-Reference-Format}
\balance
\bibliography{sydes}

\appendix

\section{Limitation}\label{app:limitation}

	Although we thoroughly explore non-verbal cues in images through a masked image modeling and our proposed method demonstrated effectiveness on the MSED dataset, several limitations remain: 
	
	\begin{enumerate}
		\item Capturing fine-grained local features is still insufficient. As shown in Section~\ref{subsec452}, reconstructions of small but semantically important region such as faces are limited, indicating that there is still room for improvement in perceiving and reconstructing fine-grained regions. 
		\item The exploitation of textual information could be further enhanced. Textual descriptions often obtain abundant aspect-level expressions (e.g., opinion words). Effectively leveraging such terms, integrating and aligning them with visual information represents a promising direction for future research. 
		\item The current method focuses only on image-text pairs. Real-world multimodal data involve additional modalities (e.g., video and audio), which provide richer non-verbal cues. Extending our method to incorporate more modalities while balancing computational cost and performance constitutes an important direction for further investigation.
	\end{enumerate}

		\begin{table}[bp!]
			\centering
			\caption{Data statistics of MSED dataset}
			\label{tab:msed_stats}
			\renewcommand{\arraystretch}{0.55}
			\resizebox{\columnwidth}{!}{%
				\begin{NiceTabular}{c | c | c c c}
					\toprule
					\textbf{Task} & \textbf{Label} & \textbf{Train} & \textbf{Validation} & \textbf{Test} \\
					\midrule
					\multirow{7}{*}{Desire Understanding} 
					& Vengeance    & 277  & 39  & 75  \\
					& Curiosity    & 634  & 118 & 213 \\
					& Social-contact & 437  & 59  & 138 \\
					& Family       & 873  & 152 & 288 \\
					& Tranquility  & 245  & 39  & 87  \\
					& Romance      & 692  & 101 & 210 \\
					& None         & 2,969 & 507 & 1,031 \\
					\midrule
					\multirow{6}{*}{Emotion Recognition} 
					& Happiness & 2,524 & 419 & 860 \\
					& Sad       & 666  & 102 & 186 \\
					& Neutral   & 1,664 & 294 & 569 \\
					& Disgust   & 251  & 44  & 80  \\
					& Anger     & 523  & 78  & 172 \\
					& Fear      & 499  & 84  & 175 \\
					\midrule
					\multirow{3}{*}{Sentiment Analysis} 
					& Positive & 2,524 & 419 & 860 \\
					& Neutral  & 1,664 & 294 & 569 \\
					& Negative & 1,939 & 308 & 613 \\
					\bottomrule
				\end{NiceTabular}%
			}
		\end{table}
		
\section{Dataset and Training Details}\label{app:app1}

	\subsection{Dataset Statistics}\label{app:msed_stats}
	
		The complete, per-class detailed statistics are provided in Table~\ref{tab:msed_stats}.
	
	\subsection{Training Hyperparameters}\label{app:training}
	
		The complete training and implementation details for both pre-training and fine-tuning stages are summarized in Table~\ref{tab:training_hyperparams}.

\section{Detailed Experimental Results}\label{app:app2}

	\subsection{Contributions of Different Modalities}\label{app:compelete_diff_modalities}
	
		As shown in Table~\ref{tab:abl_complet_diff_modal}, we conduct an experiment to analyze the contribution of non-verbal cues. Our method achieves a significant improvement under the image-only setting. This indicates that our method can effectively utilize non-verbal cues from images. In contrast, performance gains are limited or even degrade under the text-only setting, further confirming that our model focuses more on visual information. In the multimodal setting, our method consistently outperforms the baseline, indicating that it enhances visual signal while preserving textual understanding. These results demonstrate that the effectiveness of our proposed method in leveraging non-verbal cues from images.

		\begin{table}[bp!]
			\centering
			\caption{Pretraining and fine-tuning hyperparameters}
			\label{tab:training_hyperparams}
			\resizebox{0.8\columnwidth}{!}{%
				\begin{NiceTabular}{lcc}
					\toprule
					\textbf{Hyperparameter} & \textbf{Pre-training} & \textbf{Fine-tuning} \\
					\midrule
					\multicolumn{3}{l}{\textbf{Configuration}} \\
					\hspace{1em}Batch Size & 64 & 64 \\
					\hspace{1em}Epochs & 50 & 50 \\
					\hspace{1em}Optimizer & \multicolumn{2}{c}{AdamW \cite{Loshchilov2017DecoupledWD}} \\
					\hspace{1em}LR Schedule & \multicolumn{2}{c}{Cosine Decay} \\
					\hspace{1em}Weight Decay & 0.01 & 0.01 \\
					\hspace{1em}Warmup Ratio & 0.15 & 0.10 \\
					\hspace{1em}Mask Ratio & 0.75 & 0.00 \\
					\midrule
					\multicolumn{3}{l}{\textbf{Learning Rates}} \\
					\hspace{1em}Image Encoder & $5 \times 10^{-6}$ & - \\
					\hspace{1em}Text Encoder & $5 \times 10^{-5}$ & $1 \times 10^{-4}$ \\
					\hspace{1em}Image Decoder & $1 \times 10^{-4}$ & - \\
					\hspace{1em}Text Decoder & - & $2 \times 10^{-4}$\\
					\hspace{1em}MLP Classifier & - & $1 \times 10^{-4}$ \\
					\midrule
					\multicolumn{3}{l}{\textbf{Loss Weights}} \\
					\hspace{2em}$\lambda_{\mathrm{rec}}$ & 1.0 & - \\
					\hspace{2em}$\lambda_{\mathrm{si}}$ & 0.5 & - \\
					\hspace{2em}$\lambda_{\mathrm{dc}}$ & 0.025 & - \\
					\hspace{2em}$\lambda_{\mathrm{itc}}$ & 0.5 & 0.4 \\
					\hspace{2em}$\lambda_{\mathrm{cls}}$ & - & 1.0 \\
					\bottomrule
				\end{NiceTabular}%
			}
		\end{table}

	\subsection{Contributions of Loss Functions} \label{app:compelete_diff_losses}

		To provide a more comprehensive contribution analysis of each loss function in our proposed framework, we conduct extensive ablation studies as shown in Table~\ref{tab:loss_ablation_comprehensive}. The experiments systematically investigate the necessity and effectiveness of each loss function in both pre-training and fine-tuning stages.

		\subsubsection{Experimental Design}
		
			Our ablation study includes the following configurations:

			\begin{itemize}
				\item \textbf{Baseline (Row 1)}: Training with only classification loss $\mathcal{L}_{\text{cls}}$ in fine-tuning stage, without any pre-training.
				
				\item \textbf{Individual Loss Analysis (Rows 2-5)}: We first examine individual components separately:
				\begin{itemize}
					\item Only $\mathcal{L}_{\text{rec}}$ in pre-training (Row 2-3)
					\item Only $\mathcal{L}_{\text{itc}}$ in pre-training (Row 4-5)
				\end{itemize}
				
				\item \textbf{Loss Combinations (Rows 6-8)}: We investigate the interaction between different losses:
				\begin{itemize}
					\item $\mathcal{L}_{\text{rec}}$ + $\mathcal{L}_{\text{itc}}$ combination (Row 6)
					\item $\mathcal{L}_{\text{rec}}$ + $\mathcal{L}_{\text{si}}$ + $\mathcal{L}_{\text{dc}}$ combination (Row 7)
					\item $\mathcal{L}_{\text{rec}}$ + $\mathcal{L}_{\text{itc}}$ + $\mathcal{L}_{\text{si}}$ combination (Row 8)
				\end{itemize}
				
				\item \textbf{Stage-specific Ablations (Row 9)}: We examine the importance of $\mathcal{L}_{\text{itc}}$ in fine-tuning stage.
				
				\item \textbf{Complete Model (Row 10)}: Our complete framework with all losses.
			\end{itemize}

		\subsubsection{Key Findings}

			From the comprehensive results, we observe several important patterns:

			\noindent \textbf{Effectiveness of $\mathcal{L}_{\text{itc}}$.} The image-text contrastive loss plays a crucial role in both stages. When $\mathcal{L}_{\text{itc}}$ is removed from pre-training (Row 7), performance drops significantly across all tasks. Similarly, removing $\mathcal{L}_{\text{itc}}$ from fine-tuning (Row 9) also leads to notable performance degradation, confirming its necessity in both stages.
			
			\noindent \textbf{Synergy between Losses.} The combination of $\mathcal{L}_{\text{rec}}$ and $\mathcal{L}_{\text{itc}}$ (Row 6) shows substantial improvement over using either component alone (Rows 2-5), demonstrating their complementary nature. Adding $\mathcal{L}_{\text{si}}$ (Row 8) further improves Desire Understanding performance, validating the importance of structural information.
			
			\noindent \textbf{Progressive Improvement}. Performance improves progressively as we add more constrained losses, with the complete model achieving the best results across all metrics.
			
			\noindent \textbf{Cross-Task Consistency}. The ablation patterns are consistent across all three tasks, suggesting that our loss components provide general benefits rather than task-specific optimizations.
	
			\begin{table*}[htb]
				\centering
				\caption{Complete experimental results on the contributions of different Modalities. We report Precision (P), Recall (R), F1-score (F1), and Accuracy (Acc). Under the \textit{Image-Only} setting, the input text is an empty string; under the \textit{Text-Only} setting, the input image is an all-one tensor. The baseline is SyDES-B.}
				\label{tab:abl_complet_diff_modal}
				\resizebox{1.00\textwidth}{!}{%
					\begin{NiceTabular}{c | cc | *{4}{c} *{4}{c} *{4}{c}}
						\toprule
						\multirow{2}{*}{\textbf{Methods}} & 
						\multicolumn{2}{c|}{\textbf{Modal}} & 
						\multicolumn{4}{c}{\textbf{Desire Understanding}} & 
						\multicolumn{4}{c}{\textbf{Emotion Recognition}} & 
						\multicolumn{4}{c}{\textbf{Sentiment Analysis}} \\
						\cmidrule(lr){2-3} \cmidrule(lr){4-7} \cmidrule(lr){8-11} \cmidrule(l){9-15}
						& Text & Image & P & R & F1 & Acc & P & R & F1 & Acc & P & R & F1 & Acc \\
						\midrule
						\multirow{3}{*}{SyDES-B} & $\times$ & $\checkmark$ & 49.07 & 28.92 & 25.16 & 50.20 & 69.15 & 63.25 & 62.52 & 72.92 & 78.71 & 78.71 & 78.02 & 79.04 \\
						& $\checkmark$ & $\times$ & 40.32 & 29.05 & 29.00 & 58.08 & 68.01 & 65.06 & 62.35 & 62.14 & 77.86 & 74.01 & 72.89 & 71.79 \\
						& \cellcolor{lightgray!30}$\checkmark$ & \cellcolor{lightgray!30}$\checkmark$ & \cellcolor{lightgray!30}80.77 & \cellcolor{lightgray!30}67.43 & \cellcolor{lightgray!30}72.27 & \cellcolor{lightgray!30}80.51 & \cellcolor{lightgray!30}81.04 & \cellcolor{lightgray!30}80.66 & \cellcolor{lightgray!30}80.80 & \cellcolor{lightgray!30}82.61 & \cellcolor{lightgray!30}89.09 & \cellcolor{lightgray!30}85.58 & \cellcolor{lightgray!30}86.50 & \cellcolor{lightgray!30}87.12 \\
						\midrule
						\multirow{3}{*}{SyDES} & $\times$ & $\checkmark$ & 75.61 & 63.35 & 65.62 & 77.91 & 72.47 & 69.74 & 70.71 & 76.10 & 83.81 & 81.07 & 82.02 & 82.52 \\
						& $\checkmark$ & $\times$ & 43.96 & 35.82 & 37.24 & 58.67 & 65.08 & 48.54 & 49.53 & 62.49 & 78.30 & 74.02 & 72.46 & 71.35 \\
						& \cellcolor{lightgray!30}$\checkmark$ & \cellcolor{lightgray!30}$\checkmark$ & \cellcolor{lightgray!20}{84.09} & \cellcolor{lightgray!30}{84.07} & \cellcolor{lightgray!30}{84.02} & \cellcolor{lightgray!30}{88.32} & \cellcolor{lightgray!30}{84.92} & \cellcolor{lightgray!30}{84.81} & \cellcolor{lightgray!30}{84.74} & \cellcolor{lightgray!30}{85.96} & \cellcolor{lightgray!30}{89.28} & \cellcolor{lightgray!30}{89.13} & \cellcolor{lightgray!30}{89.19} & \cellcolor{lightgray!30}{89.37} \\
						\bottomrule
					\end{NiceTabular}%
				}
			\end{table*}
				
			\begin{table*}[htb]
				\centering
				\caption{Comprehensive ablation study on the loss functions in the two-stage training}
				\label{tab:loss_ablation_comprehensive}
				\resizebox{1.00\textwidth}{!}{%
					\begin{NiceTabular}{cccc | cc | *{4}{c} *{4}{c} *{4}{c}}
						\toprule
						\multicolumn{4}{c}{\textbf{Pre-training}} & 
						\multicolumn{2}{c|}{\textbf{Fine-tuning}} & 
						\multicolumn{4}{c}{\textbf{Desire Understanding}} & 
						\multicolumn{4}{c}{\textbf{Emotion Recognition}} & 
						\multicolumn{4}{c}{\textbf{Sentiment Analysis}} \\
						\cmidrule(lr){1-4} \cmidrule(lr){5-6} \cmidrule(lr){7-10} \cmidrule(lr){11-14} \cmidrule(l){15-18}
						$\mathcal{L}_{\text{rec}}$ & $\mathcal{L}_{\text{itc}}$ & $\mathcal{L}_{\text{si}}$ & $\mathcal{L}_{\text{dc}}$ & $\mathcal{L}_{\text{itc}}$ & $\mathcal{L}_{\text{cls}}$ & P & R & F1 & Acc & P & R & F1 & Acc & P & R & F1 & Acc \\
						\midrule
						
						\rowcolor{lightgray!20}
						$\times$ & $\times$ & $\times$ & $\times$ & $\times$ & $\checkmark$ & 80.77 & 67.43 & 72.27 & 80.51 & 81.04 & 80.66 & 80.80 & 82.61 & 89.09 & 85.58 & 86.50 & 87.12 \\
						
						\rowcolor{yellow!10}
						$\checkmark$ & $\times$ & $\times$ & $\times$ & $\times$ & $\checkmark$ & 73.01 & 72.41 & 72.63 & 79.33 & 75.83 & 73.55 & 73.99 & 75.86 & 80.00 & 80.90 & 80.33 & 80.51 \\
						
						\rowcolor{yellow!10}
						$\checkmark$ & $\times$ & $\times$ & $\times$ & $\checkmark$ & $\checkmark$ & 76.17 & 72.56 & 74.12 & 81.15 & 76.78 & 75.13 & 75.82 & 77.62 & 80.57 & 81.19 & 80.79 & 80.80 \\
						
						\rowcolor{yellow!10}
						$\times$ & $\checkmark$ & $\times$ & $\times$ & $\times$ & $\checkmark$ & 78.58 & 75.97 & 76.76 & 82.71 & 82.02 & 81.65 & 81.82 & 83.38 & 86.54 & 86.90 & 86.70 & 86.86 \\
						
						\rowcolor{yellow!10}
						$\times$ & $\checkmark$ & $\times$ & $\times$ & $\checkmark$ & $\checkmark$ & 80.26 & 81.95 & 80.96 & 84.19 & 83.09 & 82.15 & 82.57 & 84.43 & 86.41 & 86.89 & 86.61 & 86.64 \\
						
						\rowcolor{red!5}
						$\checkmark$ & $\checkmark$ & $\times$ & $\times$ & $\checkmark$ & $\checkmark$ & 81.43 & 81.77 & 81.44 & 86.19 & 84.85 & 84.95 & 84.66 & 85.21 & 89.26 & 88.80 & 88.99 & 89.18 \\
						
						\rowcolor{red!5}
						$\checkmark$ & $\times$ & $\checkmark$ & $\checkmark$ & $\checkmark$ & $\checkmark$ & 75.45 & 73.52 & 74.39 & 80.31 & 76.49 & 74.05 & 75.07 & 77.23 & 80.51 & 80.95 & 80.71 & 80.85 \\
						
						\rowcolor{red!5}
						$\checkmark$ & $\checkmark$ & $\checkmark$ & $\times$ & $\checkmark$ & $\checkmark$ & 82.28 & 82.33 & 82.24 & 86.78 & 84.17 & 81.91 & 82.88 & 84.53 & 87.45 & 87.66 & 87.55 & 87.71 \\
						
						\rowcolor{blue!5}
						$\checkmark$ & $\checkmark$ & $\checkmark$ & $\checkmark$ & $\times$ & $\checkmark$ & 78.55 & 75.89 & 77.09 & 83.74 & 84.85 & 81.75 & 83.21 & 84.43 & 86.46 & 85.93 & 86.16 & 86.44 \\
						
						\rowcolor{green!10}
						\textbf{$\checkmark$} & \textbf{$\checkmark$} & \textbf{$\checkmark$} & \textbf{$\checkmark$} & \textbf{$\checkmark$} & \textbf{$\checkmark$} & \textbf{84.09} & \textbf{84.07} & \textbf{84.02} & \textbf{88.32} & \textbf{84.92} & \textbf{84.81} & \textbf{84.74} & \textbf{85.96} & \textbf{89.28} & \textbf{89.13} & \textbf{89.19} & \textbf{89.37} \\
						
						\bottomrule
					\end{NiceTabular}%
				}
			\end{table*}

\end{document}